\documentclass{article} 
\usepackage{iclr2017_workshop,times}
\usepackage{url}
\usepackage{amsmath}
\usepackage{amsfonts}
\usepackage{amssymb}
\usepackage{graphicx} 
\usepackage{algorithm}
\usepackage{algorithmic}
\usepackage{appendix}
\usepackage{booktabs}
\usepackage{bbm}
\usepackage{caption}
\usepackage{subcaption}
\usepackage{natbib}
\usepackage[colorinlistoftodos]{todonotes}

\title{Precise Recovery of Latent Vectors from\\ Generative Adversarial Networks}

\author{Zachary C Lipton \thanks{http://zacklipton.com} \\
CSE Department\\
University of California, San Diego\\
\texttt{zlipton@cs.ucsd.edu}\\ 
\And
Subarna Tripathi \thanks{http://acsweb.ucsd.edu/~stripath/} \\
ECE Department\\
University of California, San Diego\\
\texttt{stripathi@eng.ucsd.edu} \\
}

\begin{document}

\maketitle

\begin{abstract}
Generative adversarial networks (GANs)
transform latent vectors into visually plausible images. 
It is generally thought that the original GAN formulation gives no out-of-the-box method to reverse the mapping, 
projecting images back into latent space. 
We introduce a simple, gradient-based technique called \emph{stochastic clipping}. 
In experiments, for images generated by the GAN, 
we precisely recover their latent vector pre-images $100\%$ of the time.
Additional experiments demonstrate that this method is robust to noise.
Finally, we show that even for unseen images, our method appears to recover unique encodings.

\end{abstract}

\section{Introduction}
\label{sec:introduction}
Deep convolutional neural networks (CNNs) are now standard tools for machine learning practitioners.
Currently, they outperform all other computer vision techniques for discriminative learning problems 
including image classification and object detection. 
Generative adversarial networks (GANs) \citep{goodfellow2014generative,radford2015unsupervised} 
adapt deterministic deep neural networks to the task of generative modeling.

GANs consist of a \emph{generator} and a \emph{discriminator}.
The generator maps samples from a low-dimensional latent space onto the space of images.  
The discriminator tries to distinguish between images produced by the generator and real images.
During training, the generator tries to \emph{fool} the discriminator.
After training, researchers typically discard the discriminator. 
Images can then be generated by drawing samples from the latent space and passing them through the generator.


While the generative capacity of GANs is well-known, how best to perform the reverse mapping (from image space to latent space) remains an open research problem.
\citet{donahue2016adversarial} suggests an extension to GAN in which a third model explicitly learns the reverse mapping.
\cite{creswell2016inverting} suggest that inverting the generator is difficult, noting that, in principle, a single image $\phi({\boldsymbol{z}})$ may map to multiple latent vectors $\boldsymbol{z}$. 
They propose a gradient-based approach to recover latent vectors
and evaluate the process on the reconstruction error 
in image space. 

We reconstruct latent vectors by performing gradient descent over the components of the latent representations and introduce a new technique called \emph{stochastic clipping}. 
To our knowledge, this is the first empirical demonstration 
that DCGANS can be inverted to arbitrary precision.
Moreover, we demonstrate that these reconstructions are robust to added noise.
After adding small amounts of Gaussian noise to images,
we nonetheless recover the latent vector $\boldsymbol{z}$ with little loss of fidelity. 

In this research, we also seek insight regarding the optimizations over neural network loss surfaces. We seek answers to the questions: 
(i) Will the optimization achieve the globally minimal loss of $0$ or get stuck in sub-optimal critical points?
(ii) Will the optimization recover precisely the same input every time?
Over $1000$ experiments, 
we find that on a pre-trained DCGAN network, 
gradient descent with \emph{stochastic clipping}  
recovers the true latent vector
$100\%$ of the time to arbitrary precision.

\textbf{Related Work: }
Several papers attempt gradient-based methods for inverting deep neural networks.
\citet{mahendran2015understanding} invert discriminative CNNs to understand hidden representations.
\citet{creswell2016inverting} invert the generators of GANS but do not report finding faithful reconstruction in the latent space.
We note that the task of finding pre-images for non-convex mappings has a history in computer vision dating at least as far back as \citet{bakir2004learning}.

\section{Gradient-Based Input Reconstruction and Stochastic Clipping}
\label{sec:methods}
To invert the mappings learned by the \emph{generator}, we apply the following idea. 
For a latent vector $\boldsymbol{z}$, 
we produce an image $\phi(\boldsymbol{z})$. 
We then initialize a new, random vector $\boldsymbol{z'}$ 
of the same shape as $\boldsymbol{z}$.
This new vector $\boldsymbol{z'}$ 
maps to a corresponding image $\phi(\boldsymbol{z'})$.
In order to reverse engineer the input $\boldsymbol{z}$, 
we successively update the components of $\boldsymbol{z'}$ in order to push the representation $\phi(\boldsymbol{z'})$ closer to the original image $\phi(\boldsymbol{z})$. 
In our experiments we minimize the $L_2$ norm, yielding the following optimization problem:
$$\min_{\boldsymbol{z'}} || \phi(\boldsymbol{z}) - \phi(\boldsymbol{z'})||^2_2.$$
We optimize over $\boldsymbol{z'}$ by gradient descent, performing the update $\boldsymbol{z'} \gets \boldsymbol{z'} - \eta \nabla_{\boldsymbol{z'}} || \phi(\boldsymbol{z}) - \phi(\boldsymbol{z'})||^2_2 $ until some convergence criteria is met.
The learning rate $\eta$ is attenuated over time. 

Note that this optimization has a global minima of $0$. 
However, we do not know if the solution that achieves this global minimum is unique. 
Moreover, this optimization is non-convex,
and thus we know of no theoretical reason that this optimization should precisely recover the original vector. 

In many cases, we know that the original input 
comes from a bounded domain. 
For DCGANS, all latent vectors are sampled uniformly from the $[-1,1]^{100}$ hyper-cube. 
To enforce this constraint, 
we apply the modified optimization
$$\boldsymbol{z'} \gets \text{clip}(\boldsymbol{z'} - \alpha \nabla_{\boldsymbol{z}'} || \phi(\boldsymbol{z}) - \phi(\boldsymbol{z'})||^2_2 ).$$
With \emph{standard clipping},
we replace components that are too large 
with the maximum allowed value 
and components that are too small 
with the minimum allowed value.
Standard clipping precisely recovers a large fraction of vectors $\boldsymbol{z}$. 

For the failure cases, we noticed that the reconstructions ${z'}$ had some components stuck at either $-1$ or $1$. 
Because $\boldsymbol{z} \sim \text{uniform}([-1,1]^{100})$, we know that the probability that a component should lie right at the boundary is close to zero.
To prevent these reconstructions from getting stuck, we introduce a heuristic technique called \emph{stochastic clipping}.
When using stochastic clipping, instead of setting components to $-1$ or $1$, we reassign the clipped components uniformly at random in the allowed range. 
While this can't guard against an interior local minima, it helps if the only local minima contain components stuck at the boundary.


\section{Experiments}
\label{sec:experiments}
\begin{figure*}[t!]
	\centering
    \begin{subfigure}[b]{0.25\textwidth}
  		\centering
		\includegraphics[width=\linewidth]{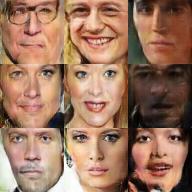}
        \caption{Imposter initialization}
\end{subfigure}~
\begin{subfigure}[b]{0.25\textwidth} 
  		\centering
		\includegraphics[width=\linewidth]{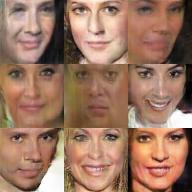}
		\caption{After $100$ iterations}
\end{subfigure}~
\begin{subfigure}[b]{0.25\textwidth} 
  		\centering
		\includegraphics[width=\linewidth]{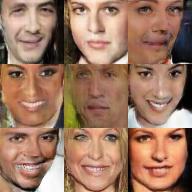}
		\caption{After $20k$ iterations}
\end{subfigure}~
\begin{subfigure}[b]{0.25\textwidth} 
  		\centering
		\includegraphics[width=\linewidth]{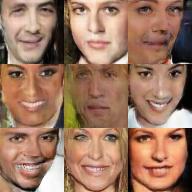}
		\caption{Original Images}
\end{subfigure}~
 \caption{Reconstruction visualizations. 
}
\label{fig:reconstruction-visualizations1}
\end{figure*}

We now summarize our experimental findings. All experiments are conducted with DCGANs as described by \citet{radford2015unsupervised} and re-implemented in Tensorflow by \citet{amos2016image}. 
First, we visualize the reconstruction process showing $\phi(\boldsymbol{z}')$ after initialization, $100$ iterations, and $20$k iterations (Figure \ref{fig:reconstruction-visualizations1}). The reconstruction ($\boldsymbol{z}'$) produces an image indistinguishable from the original.

Next, we consider the fidelity of reconstruction after $100k$ updates. 
In Table \ref{tab:reconstruction-results}, we show that even with conservative thresholds for determining reconstruction success, stochastic thresholding recovers $100\%$ percent of latent vectors. 
We evaluate these numbers using $1000$ examples.

\begin{table*}[hbt!]
\centering
\footnotesize
\begin{tabular}{lllll} 
& \multicolumn{4}{c}{\textbf{Accuracy Threshold}}
\\
\toprule
\textbf{\textbf{Technique}} & $\boldsymbol{10^{-4}}$ & $\boldsymbol{10^{-3}}$ & $\boldsymbol{10^{-2}}$ & $\boldsymbol{10^{-1}}$ 
\\
\midrule
\textbf{No Clipping} & $.98$ &$.986$ &$.998$ &$1.0$
\\
\textbf{Standard Clipping}   & $.984$ &$.99$ & $1.0$ &$1.0$
\\
\textbf{Stochastic Clipping}  & $1.0$ & $1.0$ &$1.0 $& $1.0$ 
\\
\bottomrule
\end{tabular} 
\caption{
Percentage of successful reconstructions $||\boldsymbol{z} -\boldsymbol{z}'||^2_2 / 100 < \epsilon$ for various thresholds $\epsilon$ out of $1000$ trials. 
}
\label{tab:reconstruction-results}
\end{table*}

\begin{figure*}[t!]
	\centering
    \begin{subfigure}[b]{0.5\textwidth}
  		\centering
		\includegraphics[scale=.5]{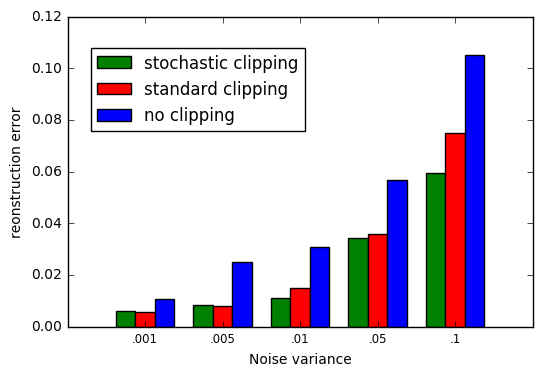}
        \caption{$z$-space reconstruction with noise}
\end{subfigure}~
\begin{subfigure}[b]{0.5\textwidth} 
  		\centering
		\includegraphics[scale=.45]{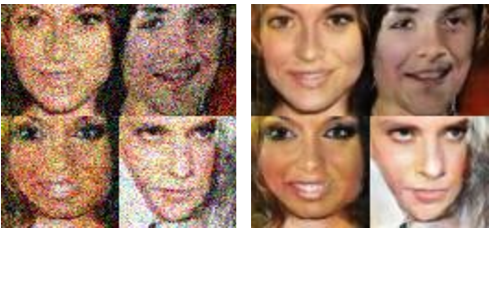}
		\caption{$\phi(\boldsymbol{z}) + \boldsymbol{\eta}$ and $\phi(\boldsymbol{z'})$ 
        for noise variance .1 and recovery by stochastic clipping}
\end{subfigure}
 \caption{$\boldsymbol{z}$-space reconstruction error grows proportional to added noise. 
\emph{Stochastic clipping} appears more robust to noise than other methods. Because pixel values are scaled $[-1,1]$, noise variance of $.1$ corresponds to a standard deviation of $39.5$ pixel values.
}
\label{fig:robustness-plots}
\end{figure*}

We then consider the robustness of these reconstructions to noise.
We apply Gaussian white noise $\boldsymbol{\eta}$, attempting to reconstruct $\boldsymbol{z}$ from $\phi(\boldsymbol{z}) + \boldsymbol{\eta}$.
Our experiments show that even for substantial levels of noise, the reconstruction error in $\boldsymbol{z}$-space is low and appears to grow proportionally to the added noise (Figure \ref{fig:robustness-plots}).


Finally, we ask whether for unseen images, the recovered vector is always the same. To determine the consistency of the recovered vector, we recover 1000 vectors for the same image and plot the average pair-wise distance between reconstructions.  
\begin{table*}[hbt!]
\centering
\footnotesize
\begin{tabular}{llll|l}
\\
\toprule
\textbf{\textbf{Clipping}} &  \textbf{None} & 
\textbf{Standard} & \textbf{Stochastic} & \textbf{Baseline}
\\
\midrule
$\mathbbm{E}\left(\frac{||\boldsymbol{z}_i' - \boldsymbol{z}_j'||}{100}\right)$ &$2.03\text{e-4}$ &$1.36\text{e-4}$ &$4.94\text{e-5}$& $.0815$
\\
\bottomrule
\end{tabular} 
\caption{Average pairwise distance of recovered vectors for unseen images. The baseline score is the average distance between random vectors sampled randomly from the latent space. 
}
\label{tab:unseen-images}
\vspace{-2mm}
\end{table*}



\section{Conclusions}
\label{sec:conclusion}
We show that GAN generators can, in practice, be inverted to arbitrary precision. 
These inversions are robust to noise and the inversions appear unique even for unseen images.
Stochastic clipping is both more accurate and more robust than standard clipping. 
We suspect that stochastic clipping should also give better and more robust reconstructions of images from discriminative CNN reconstructions, leaving these experiments to future work.

\bibliography{iclr2017_workshop}
\bibliographystyle{iclr2017_workshop}

\end{document}